\pdfoutput=1
\documentclass[11pt]{article}

\usepackage[]{emnlp2021}

\usepackage{times}
\usepackage{latexsym}

\usepackage[T1]{fontenc}
\usepackage[utf8]{inputenc}

\usepackage{microtype}

\usepackage{url} % added by me
\usepackage{graphicx} % added by me
\usepackage{subcaption} % added by me
\usepackage{comment} % added by me
\usepackage{makecell} % added by me

\newcommand\blfootnote[1]{%
  \begingroup
  \renewcommand\thefootnote{}\footnote{#1}%
  \addtocounter{footnote}{-1}%
  \endgroup
}

\title{How much pretraining data do language models need to learn syntax?}

\author{
  Laura Pérez-Mayos\textsuperscript{*1}, Miguel Ballesteros\textsuperscript{2}, Leo Wanner\textsuperscript{3,1} \\[5pt]
  \textsuperscript{1} TALN Research Group, Pompeu Fabra University, Barcelona, Spain \\
  \textsuperscript{2} Amazon AI \\
  \textsuperscript{3} Catalan Institute for Research and Advanced Studies (ICREA), Barcelona, Spain \\[5pt]
  \texttt{\{laura.perezm$\mid$leo.wanner\}@upf.edu} \\
  \texttt{ballemig@amazon.com}
}

\begin{document}
\maketitle

\begin{abstract}

Transformers-based pretrained language models achieve outstanding results in many well-known NLU benchmarks. However, while pretraining methods are very convenient, they are expensive in terms of time and resources. This calls for a study of the impact of pretraining data size on the knowledge of the models.  We explore this impact on the syntactic capabilities of RoBERTa, using models trained on incremental sizes of raw text data. First, we use syntactic structural probes to determine whether models pretrained on more data encode a higher amount of syntactic information. Second, we perform a targeted syntactic evaluation to analyze the impact of pretraining data size on the syntactic generalization performance of the models. Third, we compare the performance of the different models on three downstream applications: part-of-speech tagging, dependency parsing and paraphrase identification. We complement our study with an analysis of the cost-benefit trade-off of training such models. Our experiments show that while models pretrained on more data encode more syntactic knowledge and perform better on downstream applications, they do not always offer a better performance across the different syntactic phenomena and come at a higher financial and environmental cost.

\end{abstract}

\section{Introduction}
\label{sec:intro}

\blfootnote{$\ast$ Work partially done during internship at Amazon AI.}

The use of unsupervised pretrained language models in the context of supervised tasks has become a widely spread practice in NLP, with Transformer-based models such as BERT \citep{devlin-etal-2019-bert} and RoBERTa \citep{liu2019roberta} achieving outstanding results in many well-known Natural Language Understanding benchmarks such as GLUE \citep{wang-etal-2018-glue} and SQuAD \cite{rajpurkar-etal-2018-know}. Consequently, several studies investigate the types of knowledge learned by BERT, how and where this knowledge is represented and what the best methods to improve it are; see, e.g., \cite{rogers-etal-2020-primer}. There is evidence that, among other information (e.g., part-of-speech, syntactic chunks and roles \citep{tenney2018what, lin-etal-2019-open, belinkov-etal-2017-neural},  morphology in general \citep{peters-etal-2018-deep}, or sentence length \citep{adi2016fine}), BERT representations implicitly embed entire syntax trees  \citep{hewitt2019structural}.

Language models are traditionally assessed by information-theoretical metrics such as perplexity, i.e., the probability of predicting a word in its context. The general wisdom is that the more pretraining data a model is fed, the lower its perplexity gets. However, large volumes of pretraining data are not always available and pretraining is costly, such that the following questions need to be answered: {\bf (i)} Do we always need models pretrained on internet-scale corpora? {\bf (ii)} As the models are pretrained on more data, and their perplexity improves, do they encode more syntactic information and offer a better syntactic generalization? {\bf (iii)} Do the models with more pretraining perform better when applied in downstream tasks? To address these questions, we explore the relation between the size of the pretraining data and the syntactic capabilities of RoBERTa by means of the MiniBERTas models, a set of 12 RoBERTa models pretrained from scratch by \citet{warstadt-etal-2020-learning} on quantities of data ranging from 1M to 1B words. In particular:
 
 \begin{itemize}
     \vspace*{-0.1cm}
     \item We use the syntactic structural probes from \citet{hewitt2019structural} to determine whether those models pretrained on more data encode a higher amount of syntactic information than those trained on less data; 
     \vspace*{-0.2cm}
     \item We perform a targeted syntactic evaluation to analyze the generalization performance of the different models using SyntaxGym \citep{gauthier-etal-2020-syntaxgym} and the syntactic tests presented in \citep{hu-etal-2020-systematic};
     \vspace*{-0.2cm}
     \item We compare the performance of the different models on two morpho-syntactic tasks (PoS tagging and  dependency parsing), and a non-syntactic task (paraphrase identification);
     \vspace*{-0.2cm}
     \item We conduct a cost-benefit trade-off analysis \cite{strubell-etal-2019-energy,bhattacharjee-etal-2020-bert} of the models training.
     \vspace*{-0.1cm}
\end{itemize}

We observe that models pretrained on more data encode a higher amount of syntax according to  \citet{hewitt2019structural}'s metrics, but do not always lead to a better syntactic generalization. Indeed, we find that models pretrained on less data perform equally good or even better than those pretrained on more data on 3 out of 6 syntactic test suites. When applied to  downstream tasks, the models pretrained on more data perform generally better. However, the analysis of the trade-off between the cost of training a model and its performance shows that small performance gains come at a high economical and environmental cost that should be considered when developing new models.

In what follows, Section \ref{sec:related_work} provides some background on the syntactic assessment of language models, model costs, and the  works related to ours. Section \ref{sec:experimental_setup} describes our experimental setup, introducing the MiniBERTas models and the syntactic tests as well as the downstream applications we explore. Section \ref{sec:results} presents the outcome of our experiments. Section \ref{sec:cost_benefit} offers a cost-benefit analysis of the pretraining of the different models, and Section \ref{sec:conclusions} summarizes the implications that our work has for the use of pretrained language models.

\section{Background}
\label{sec:related_work}

\subsection{Syntactic assessment of language models}
\label{subsec:capturing_syntactic_knowledge}

The targeted syntactic evaluation incorporates methods from psycholinguistic experiments, focusing on highly specific measures of language modeling performance and allowing to distinguish models with human-like representations of syntactic structure \citep{linzen-etal-2016-assessing,lau2017grammaticality,gulordava-etal-2018-colorless,marvin-linzen-2018-targeted,futrell-etal-2019-neural}. Regarding the evaluation of modern language models, \citet{warstadt-etal-2020-blimp} present a challenge set that isolates specific phenomena in syntax, morphology, and semantics, finding that state-of-the-art models struggle with some subtle semantic and syntactic phenomena, such as negative polarity items and extraction islands. \citet{hu-etal-2020-systematic} test 20 model type combinations and data sizes on 34 English syntactic test suites, finding substantial differences in syntactic generalization performance by model architecture.

Supervised probing models have also been used to test for the presence of a wide range of linguistic phenomena \citep{conneau-etal-2018-cram, liu-etal-2019-linguistic, tenney2018what, voita-titov-2020-information, elazar2020bert}, and it has been shown that entire syntax trees are embedded implicitly in BERT's vector geometry \citep{hewitt2019structural,chi-etal-2020-finding}. However, other works have criticized some probing methods, claiming that classifier probes can learn the linguistic task from training data \citep{hewitt-liang-2019-designing}, and can fail to determine whether the detected features are actually used \citep{voita-titov-2020-information, pimentel-etal-2020-information, elazar2020bert}.

\subsection{Costs of modern language models}
\label{subsec:economical_and_environmnetal_cost}

While modern language models keep growing in orders of magnitude, so do the resources necessary for their development and, consequently, also the inclusivity gap. The financial cost of the required hardware and electricity favors industry-powered research, and harms academics, students, and non-industry researchers, particularly those from emerging economies. Moreover, the training of such models is not only financially expensive, but has also a large carbon footprint. \citet{schwartz2019green} propose to report the financial cost of developing, training, and running models in order to provide baselines for the investigation of increasingly efficient methods. Along the same lines, \citet{strubell-etal-2019-energy} offer an analysis of the computation required for the research, development and hyperparameter tuning of several recently successful neural network models for NLP, and propose actionable recommendations to reduce costs and improve equity, namely 1) reporting training time and sensitivity to hyperparameters; 2) a government-funded academic compute cloud to provide equitable access to all researchers; and 3) prioritizing computationally efficient hardware and algorithms.

\subsection{Related work}
\label{subsec:effects_of_pretraining_data_size}

Several studies investigate the relation between pretraining data size and linguistic knowledge in language models. \citet{van-schijndel-etal-2019-quantity, hu-etal-2020-systematic,micheli-etal-2020-importance} find out that, given a relatively large data size (e.g., 10M words), models with less pretraining perform similarly to models with much more pretraining, concluding that model architecture plays a more important role than training data scale in yielding correct syntactic generalizations \citep{hu-etal-2020-systematic}. Complementary, \citet{raffel2020exploring} shows that performance can degrade when an unlabeled data set is small enough that it is repeated many times over the course of pretraining. In contrast, \citet{zhang2020you} argue that while relatively small datasets suffice to reliably encode most syntactic and semantic features, a much larger quantity of data is needed to master conventional NLU tasks.  This discrepancy  may be due to the difference in model architectures, pretraining techniques and the scaling and nature of the difference datasets. 

Our work differs significantly from recent works. We make use of a single architecture and data source, and focus exclusively on the syntactic capabilities of the models, offering an in-depth analysis that includes structural syntactic probing, detailed syntactic generalization, and downstream applications performance. Moreover, we also provide a cost-benefit analysis of the models.

\section{Experimental setup}
\label{sec:experimental_setup}

\subsection{The MiniBERTas models}
\label{subsec:minibertas}

The MiniBERTas are a set of 12 RoBERTa models pretrained from scratch by \citet{warstadt-etal-2020-learning} on 4 datasets containing 1B, 100M, 10M and 1M tokens, available through HuggingFace Transformers.\footnote{\url{https://huggingface.co/nyu-mll}} The datasets are sampled from Wikipedia and Smashwords -- the two datasets that make up the original pretraining dataset of BERT and that are included in the RoBERTa pretraining data. For each dataset size,  pretraining is run 25 times (10 times for 1B) with varying hyperparameter values; the three models with the lowest development set perplexity are released. For the smaller dataset, a  smaller model size is used to prevent over-fitting.
We refer to models trained on the same amount of data as a \textit{family} of models, and models inside a family as \textit{intra-family members} (e.g.,the  \textit{roberta-base-100M-1} model is a member of the \textit{roberta-base-100M} family). Table \ref{tab:hyperparameters2} offers an overview of the hyperparameters per model size.

\begin{table}
    \small
    \centering
    \begin{tabular}{llllll}
    \hline
    \textbf{Model Size}  &   \textbf{L}   &   \textbf{AH}  &   \textbf{HS}  &   \textbf{FFN} &   \textbf{P} \\
    \hline
    BASE    &   12  &   12  &   768 &   3072    &   125M \\
    MED-SMALL   &   6   &   8   &   512 &   2048    &   45M \\
    \hline
    \end{tabular}
    \caption{Hyperparameters per model sizes. AH = number of attention heads; HS = hidden size; FFN = feedforward network dimension; P = number of parameters.\\ }\label{tab:hyperparameters2}
\end{table}

\subsection{Structural probing}
\label{subsec:structural_probing_results}

\citet{hewitt2019structural}'s structural probes assess how well syntax trees are embedded in a linear transformation of the network representation space applying two different evaluations: Tree distance evaluation, in which squared L2 distance encodes the distance between words in the parse tree, and Tree depth evaluation, in which squared L2 norm encodes the depth in the parse tree.

{\bf Tree distance evaluation.} Evaluates how well the predicted distances between all pairs of words in a model reconstruct gold parse trees by computing the Undirected Unlabeled Attachment Score (\textit{UUAS}). It also computes the Spearman correlation between true and predicted distances for each word in each sentence, averaging across all sentences with lengths between 5 and 50 (we refer to as \textit{DSpr.}).

{\bf Tree depth evaluation.} Evaluates the ability of models to recreate the order of words specified by their depth in the parse tree, assessing their ability to identify the root of the sentence as the least deep word (\textit{Root \%}) and computing the Spearman correlation between the predicted and the true depth ordering, averaging across all sentences with lengths between 5 and 50 (we refer to as \textit{NSpr}). 

\subsection{Targeted syntactic evaluation}
\label{subsec:syntaxgym_test_tests}

We test the MiniBERTas on the syntactic tests assembled by \citet{hu-etal-2020-systematic}, accessible through the SyntaxGym toolkit \citep{gauthier-etal-2020-syntaxgym}. The tests are divided into 6 syntactic circuits, introduced below, based on the type of algorithm required to successfully process each construction. 

{\bf 1. Agreement:} Tests a language model for how well it predicts the number marking on English finite present tense verbs. It is composed of 3 Subject-Verb Number Agreement tests from \citet{marvin-linzen-2018-targeted}, 

{\bf 2. Center Embedding:} Tests the ability to embed a phrase in the middle of another phrase of the same type. Subject and verbs must match in a first-in-last-out order, meaning models must approximate a stack-like data-structure in order to successfully process them. The circuit is composed of 2 tests from \citet{wilcox-etal-2019-hierarchical}.

{\bf 3. Garden-Path Effects:} Measures the syntactic phenomena that result from tree structural ambiguities that give rise to locally coherent but globally implausible syntactic parses. The circuit is composed of 2 Main Verb / Reduced Relative Clause (MVRR) tests and 4 NP/Z Garden-paths (NPZ) tests, all from \citet{futrell2018rnns}.

{\bf 4. Gross Syntactic Expectation:} Tests the ability of the models to distinguish between coordinate and subordinate clauses: introducing a subordinator at the beginning of the sentence should make an ending without a second clause less probable, and should make a second clause more probable. The circuit is composed of 4 Subordination tests from \citet{futrell2018rnns}.

{\bf 5. Licensing:} Measures when a particular token must exist within the scope of an upstream licensor token. The circuit is composed of 4 Negative Polarity Item Licensing (NPI) tests and 6 Reflexive Pronoun Licensing tests, all from \citet{marvin-linzen-2018-targeted}.

{\bf 6. Long-Distance Dependencies: } Measures covariations between two tokens that span long distances in tree depth. The circuit is composed of 6 Filler-Gap Dependencies (FGD) tests from \citet{wilcox-etal-2018-rnn} and \citet{wilcox-etal-2019-structural}, and 2 Cleft tests from \citep{hu-etal-2020-systematic}.

\subsection{Encoding unidirectional context with bidirectional models} 

The tests in SyntaxGym evaluate whether models are able to assign a higher probability to grammatical and natural continuations of sentences. As RoBERTa is a bidirectional model, to be able to ask it to predict the probability of a token given the context of previous tokens we test it in a left-to-right generative setup, as done in \citep{Rongali_2020, zhu-etal-2020-dont}. More precisely, we follow \citet{wang-cho-2019-bert}'s sequential sampling procedure, which is not affected by the error that was reported in equations 1-3, related to the Non-sequential sampling procedure. To compute the probability distribution for a sentence with $N$ tokens, we start with a sequence of \textit{begin\_of\_sentence} token plus $N$ \textit{mask} tokens plus an extra \textit{mask} token to account for the \textit{end\_of\_sentence} token. For each masked position in $[1, N]$, we compute the probability distribution over the vocabulary given the left context of the original sequence, and select the probability assigned by the model to the original word. Note that this setup allows the models to know how many tokens there are in the sentences, and therefore the results are not directly comparable with those of unidirectional models, that do not have any information regarding the length of the sequence.

For example, in a Subordination test with the examples `Because the students did not like the material.' and `The students did not like the material.', we expect the model to assign a higher surprisal \citep{wilcox2019structural} to the first example, because the initial "Because" implies that the immediately following clause is not the main clause of the sentence, but instead is a subordinate that must be followed by the main clause. However, instead of finding the main clause, the model encounters a dot indicating the end of the sentence. To test whether the model has learned about subordination, we feed the models the tokens sequences [\textit{begin\_of\_sentence}, \textit{Because}, \textit{the}, \textit{students}, \textit{did}, \textit{not}, \textit{like}, \textit{the}, \textit{materials}, \textit{mask}, \textit{mask}] and [\textit{begin\_of\_sentence}, \textit{The}, \textit{students}, \textit{did}, \textit{not}, \textit{like}, \textit{the}, \textit{materials}, \textit{mask}, \textit{mask}], and compare the surprisal of the model predicting a dot `.' for the first masked position in each case.

\subsection{Downstream applications}
\label{subsec:downstream_applications}

To compare the performance of the models on downstream applications, we analyze their learning curves along the fine-tuning process on two morpho-syntactic tasks (PoS tagging and dependency parsing) and a non-syntactic task (paraphrase identification). Each task is fine-tuned for 3 epochs, with the default learning rate of 5e$^{-5}$. To mitigate the variance in performance induced by weight initialization and training data order \citep{dodge2020fine, reimers-gurevych-2017-reporting}, we repeat this process 5 times per task with different random seeds and average results.\footnote{The implementation relies in the Transformers library \citep{wolf-etal-2020-transformers} and AllenNLP \citep{gardner-etal-2018-allennlp}. For implementation details, pretrained weights and hyperparameter values, cf. the documentation of the libraries.} For {\bf PoS tagging}, we fine-tune RoBERTa with a linear layer on top of the hidden-states output for token classification.\footnote{Source: \url{https://github.com/Tarpelite/UniNLP/blob/master/examples/run_pos.py}} Dataset: Universal Dependencies Corpus for English (UD 2.5 English EWT \citep{silveira-etal-2014-gold}). For {\bf Dependency parsing}, we fine-tune a Deep Biaffine neural dependency parser \citep{dozat2016deep}. Dataset: UD 2.5 English EWT \cite{silveira-etal-2014-gold}. For {\bf Paraphrase identification},  we fine-tune RoBERTa with a linear layer on top of the pooled sentence representation.\footnote{Source: \url{https://github.com/huggingface/transformers/blob/master/examples/text-classification/run_glue.py}.} Dataset: Microsoft Research Paraphrase Corpus (MRPC) \citep{dolan-brockett-2005-automatically}.

\section{Results}
\label{sec:results}

In this section, we explore the impact of the size of pretraining data on the syntactic information encoded by RoBERTa from three different angles.

\subsection{Structural probing}
\label{subsec:structural_probing_results}

We use Hewitt and Manning's syntactic structural probes to determine whether the MiniBERTa models pretrained on more data encode a higher amount of syntactic information than those trained on less data. Following the original work, we probe layer 7 of all models, as it was shown to encode most of the syntax. Results are shown in Table \ref{tab:structural_probing}.

\textbf{Tree distance evaluation.} The models trained with more data encode better syntactic information (as measured by the probe metrics). While \textit{DSpr.} shows a less pronounced variability between family members, and smaller differences across families, \textit{UUAS} shows a higher intra-family variability and bigger differences between families. Noticeably, for the \textit{roberta-base-1B} family, there is a 7 points difference in \textit{UUAS} between model 1 and model 3, which have a difference of only 0.09 points in perplexity, highlighting the importance of training hyperparameters for the performance of the models.

\textbf{Tree depth evaluation.} As for the distance metrics, the models trained on more data show a better encoding of syntactic information. Again, the correlation shows less variability between family members and smaller differences between families, while \textit{Root \%} shows a higher intra-family variability (especially noticeable for \textit{roberta-base-10M}).

\begin{table}
    \small
    \centering
    \begin{tabular}{r}
        \\
        \textbf{Model} \\
        \hline
        1b-1 \\
        1b-2 \\
        1b-3 \\
        100m-1 \\
        100m-2 \\
        100m-3 \\
        10m-1 \\
        10m-2 \\
        10m-3 \\
        1m-1 \\
        1m-2 \\
        1m-3 \\
        \hline
    \end{tabular}
    \begin{tabular}{cc}
        \hline
        \multicolumn{2}{c}{\textbf{Tree distance eval.}} \\
        \hline
        \textbf{UUAS} & \textbf{Dspr.} \\
        \hline
         70.75  & 78.82 \\
         72.93  & 79.86 \\
         77.23  & 82.66 \\
         68.46  & 76.95 \\
         70.02  & 78.11 \\
         69.35  & 78.73 \\
         61.48  & 73.19 \\
         62.01  & 73.78 \\
         60.12  & 72.58 \\
         56.96  & 71.70 \\
         55.78  & 71.33 \\
         55.84  & 71.33 \\
        \hline
    \end{tabular}
    \begin{tabular}{cc}
        \hline
        \multicolumn{2}{c}{\textbf{Tree depth eval.}} \\
        \hline
        \textbf{Root \%} & \textbf{Nspr.} \\
        \hline
         83.92  & 85.38 \\
         83.53  & 85.92 \\
         85.13  & 86.87 \\
         81.21  & 84.06 \\
         81.25  & 84.53 \\
         79.88  & 84.59 \\
         70.88  & 81.65 \\
         70.07  & 81.89 \\
         67.14  & 80.62 \\
         57.12  & 74.16 \\
         56.56  & 74.74 \\
         57.41  & 74.46 \\
        \hline
    \end{tabular}
    \caption{Structural probing with Hewitt and Manning's syntactic structural probes. `1b-*' corresponds to the family roberta-base-1B, `100M-*' to roberta-base-100M, `10M-* to roberta-10M, and `1M-*' to roberta-med-small-1M.}
    \label{tab:structural_probing}
\end{table}

\subsection{Syntactic generalization evaluation}
\label{subsec:targeted_syntactic_evaluation_results}

We assess  the syntactic generalization performance of the different MiniBERTas models using \citet{hu-etal-2020-systematic}'s test suites (cf. Subsection \ref{subsec:syntaxgym_test_tests}) to answer the following questions: Do models pretrained on more data generalize better? Do models with lower perplexity perform better in the syntactic tests? Do models with more pretraining or better perplexity perform better in all circuits?

\begin{figure}
    \centering
    \begin{subfigure}[t]{0.49\textwidth}
        \vskip 0pt
        \includegraphics[width=\textwidth]{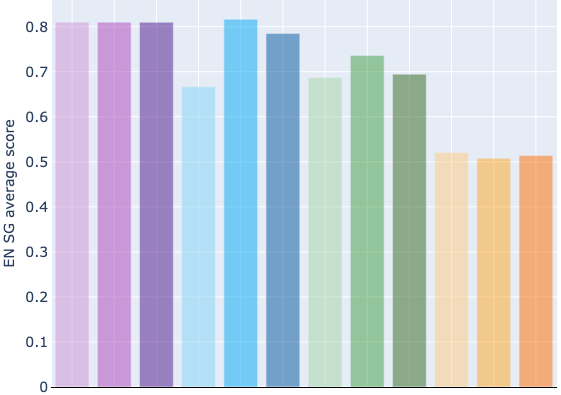}
        \label{fig:minibertas_avg_sg_score}
    \end{subfigure}
    \begin{subfigure}[t]{0.49\textwidth}
        \vskip -10pt
        \includegraphics[width=\textwidth]{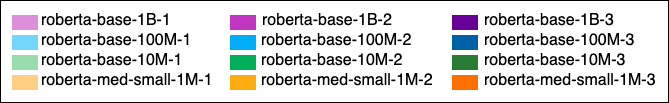}
    \end{subfigure}
    \caption{Syntactic generalization evaluation. Average SyntaxGym score.}
    \label{fig:sg_score_avg}
\end{figure}

{\bf Average SG Score.} Figure \ref{fig:sg_score_avg} shows the performance of each model averaged across all 6 circuits. We observe a variability between family members, especially for \textit{roberta-base-100M}, with a difference of 15 points between models 1 and 2. As intuitively expected, the smallest family of models, \textit{roberta-med-small-1M}, performs clearly worse than the other families. However, it is interesting to observe that more training data does not always imply better syntactic generalization: model \textit{roberta-base-100M-1} performs worse than the whole \textit{roberta-base-10M} family, and model \textit{roberta-base-100M-2} performs better than the whole \textit{roberta-base-1B} family.

\begin{figure}
    \centering
    \begin{subfigure}[t]{0.49\textwidth}
        \vskip 0pt
        \includegraphics[width=\textwidth]{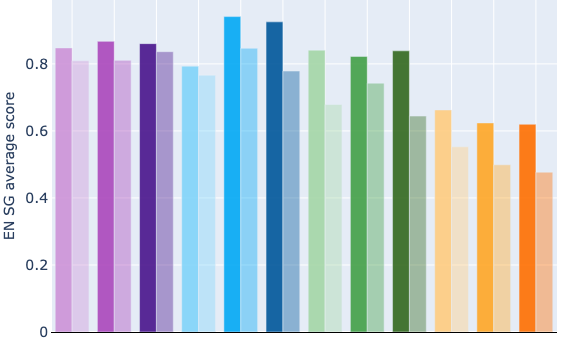}
        \label{fig:minibertas_avg_sg_score_modifiers}
    \end{subfigure}
    \begin{subfigure}[t]{0.49\textwidth}
        \vskip -10pt
        \includegraphics[width=\textwidth]{images/legend1.png}
    \end{subfigure}
    \caption{Syntactic generalization evaluation. SyntaxGym score on Center Embedding, Cleft structure, MVRR, NPZ-Verb, and NPZ-Object, without (dark bars) and with (light bars) modifiers.}
    \label{fig:sg_score_modifiers}
\end{figure}

{\bf Stability with respect to modifiers.} Five of the test suites (Center Embedding, Cleft structure, MVRR, NPZ-Verb, NPZ-Object) include tests with and without modifiers, i.e,. intervening content inserted before the critical region. These additional clauses or phrases increase the linear distance between two co-varying items, making the task more difficult, and sometimes they also include a distractor word in the middle of a syntactic dependency, which can lead the models to misinterpret the dependency. Figure \ref{fig:sg_score_modifiers} shows the models' average scores on these test suites, without modifiers (dark bars) and with modifiers (light bars), evaluating how robust each model is with respect to the intervening content. We observe that all models are affected by the presence of modifiers, but the difference is narrower for \textit{roberta-base-1b}, which offers the best stability.

\begin{figure}
    \centering
    \begin{subfigure}[t]{0.49\textwidth}
        \vskip 0pt
        \includegraphics[width=\textwidth]{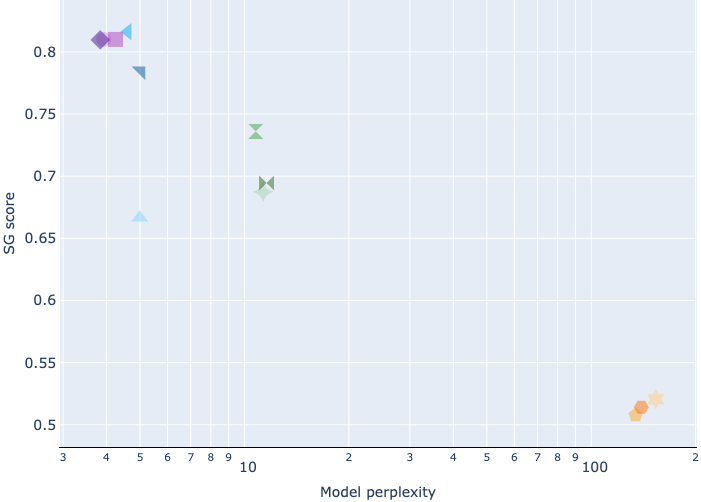}
    \end{subfigure}
    \begin{subfigure}[t]{0.49\textwidth}
        \vskip 3pt
        \includegraphics[width=\textwidth]{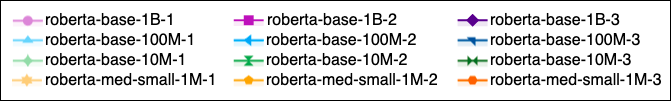}
    \end{subfigure}
    \caption{Relationship between average SyntaxGym score and model perplexity.}
    \label{fig:sg_score_vs_perplexity}
\end{figure}

{\bf Perplexity vs. SG Score.} Figure \ref{fig:sg_score_vs_perplexity} shows the relation between the average score across all circuits (\textit{SG score}) and the perplexity of the models. As previously observed in \citep{hu-etal-2020-systematic}, even though there is a (not perfect) negative correlation between the two metrics when comparing different families, when comparing points corresponding to the same family of models (with equal architecture and training data size, points of the same color in Figure \ref{fig:sg_score_vs_perplexity}), there is no clear relation between them. This suggests that both metrics capture different aspects of the knowledge of the models.

\begin{figure}
    \centering
    \begin{subfigure}[t]{0.49\textwidth}
        \vskip 0pt
        \includegraphics[width=\textwidth]{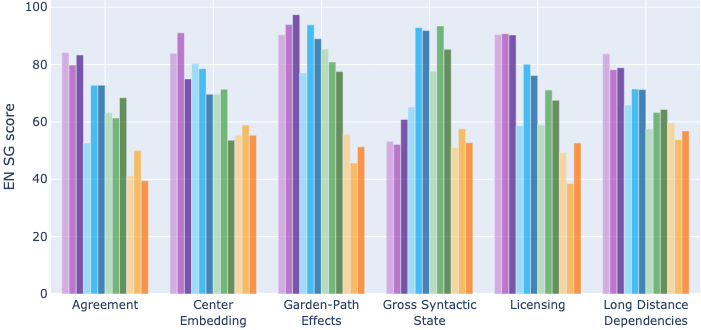}
    \end{subfigure}
    \begin{subfigure}[t]{0.49\textwidth}
        \vskip 3pt
        \includegraphics[width=\textwidth]{images/legend1.png}
    \end{subfigure}
    \caption{SyntaxGym evaluation across circuits.}
    \label{fig:minibertas_performance_across_tags_bars}
\end{figure}

{\bf Syntactic generalization of the models.} Figure \ref{fig:minibertas_performance_across_tags_bars} offers an overview of the syntactic capabilities of all the models on the different syntactic circuits. The family with more pretraining data, \textit{roberta-base-1B}, outperforms all other families in 3 out of 6 circuits, but offers a surprisingly low performance in Gross Syntactic State, clearly outperformed by  \textit{roberta-base-100M} and \textit{roberta-base-10M}, and matched by the \textit{roberta-med-small-1M}. Again, the smallest family offers the lowest performance across all circuits, with individual models outperforming isolated models of other families in Center Embedding, Gross Syntactic State and Long Distance Dependencies. There is a high variability between the scores achieved by the models of the same family in the same circuit, with the exception of \textit{roberta-base-1B} in Licensing, where all models offer a similar performance. Interestingly, there is not a single model for any family that performs best (nor worst) across all tests.\\

\subsection{Targeted downstream tasks evaluation}
\label{subsec:finetuning_minibertas_results}

\begin{figure}
    \centering
    \begin{subfigure}[t]{0.49\textwidth}
        \vskip 0pt
        \includegraphics[width=\textwidth]{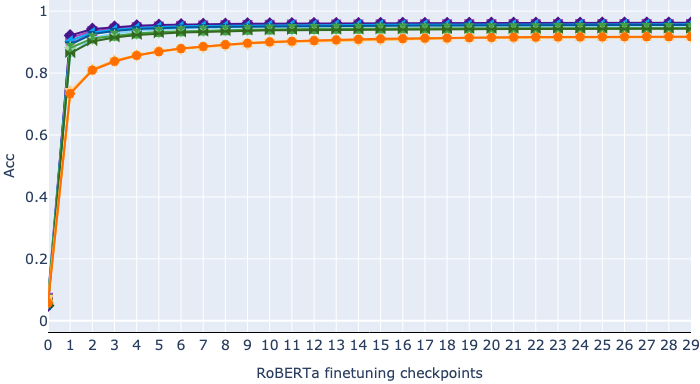}
    \end{subfigure}
    \begin{subfigure}[t]{0.49\textwidth}
        \vskip 3pt
        \includegraphics[width=\textwidth]{images/legend2.png}
    \end{subfigure}
    \caption{Targeted downstream task evaluation. PoS tagging accuracy evolution.}
    \label{fig:pos_tagging}
\end{figure}

\begin{figure}
    \centering
    \begin{subfigure}[t]{0.49\textwidth}
        \vskip 0pt
        \includegraphics[width=1\textwidth]{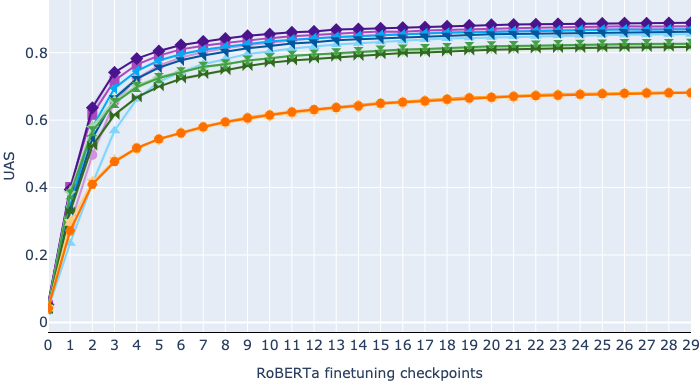}
        \label{fig:ft_dep_parsing_uas}
    \end{subfigure}
    \begin{subfigure}[t]{0.49\textwidth}
        \vskip -10pt
        \includegraphics[width=1\textwidth]{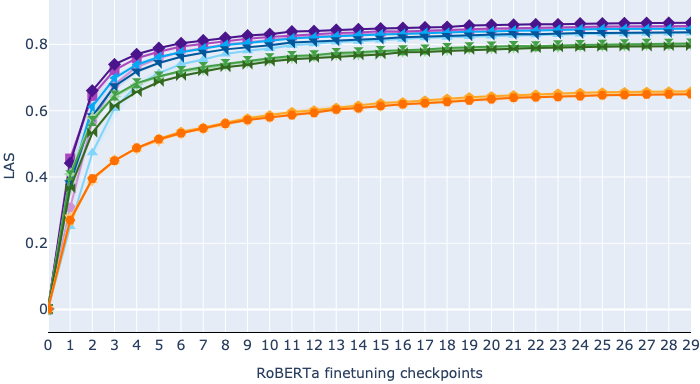}
        \label{fig:ft_dep_parsing_las}
    \end{subfigure}
    \begin{subfigure}[t]{0.49\textwidth}
        \vskip -10pt
        \includegraphics[width=\textwidth]{images/legend2.png}
    \end{subfigure}
    \caption{Targeted downstream tasks evaluation. Dependency parsing UAS and LAS evolution.}
    \label{fig:dependency_parsing}
\end{figure}

\begin{figure}
    \centering
    \begin{subfigure}[t]{0.49\textwidth}
        \vskip 0pt
        \includegraphics[width=1\textwidth]{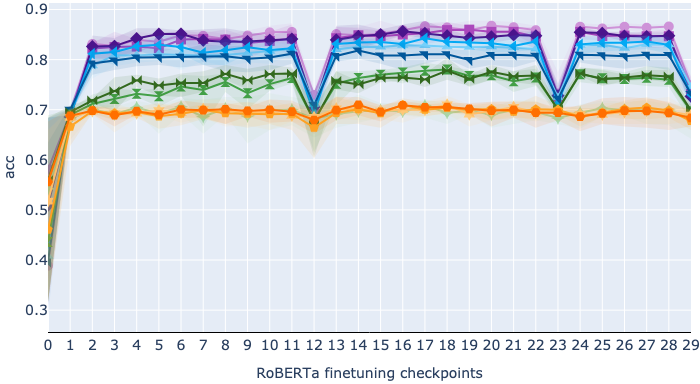}
        \label{fig:ft_paraphrasing_acc}
    \end{subfigure}
    \begin{subfigure}[t]{0.49\textwidth}
        \vskip -10pt
        \includegraphics[width=1\textwidth]{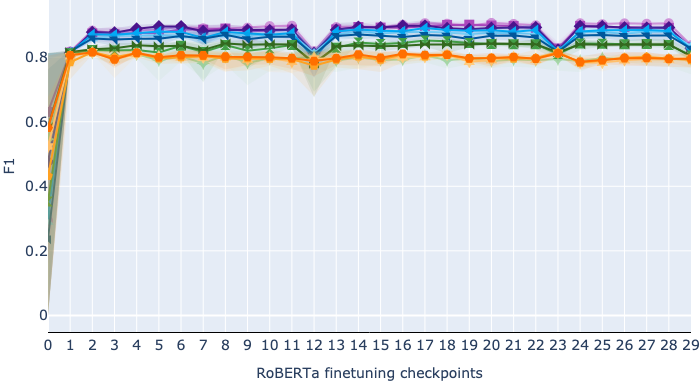}
        \label{fig:ft_paraphrasing_f1}
    \end{subfigure}
    \begin{subfigure}[t]{0.49\textwidth}
        \vskip -10pt
        \includegraphics[width=\textwidth]{images/legend2.png}
    \end{subfigure}
    \caption{Targeted downstream tasks evaluation. Paraphrase identification accuracy and F1 evolution.}
    \label{fig:paraphrase_identification}
\end{figure}

We compare the performance of the different models on three different downstream tasks: PoS  tagging (Figure \ref{fig:pos_tagging}), dependency parsing (Figure \ref{fig:dependency_parsing}) and paraphrase identification (Figures \ref{fig:paraphrase_identification}) to determine if models pretrained on more data perform better on downstream applications. We observe the same tendency for all tasks: models with more training data perform better, and the model with the smaller architecture (\textit{roberta-med-small-1M}) performs remarkably worse. Although note that while the increase of training data between families is exponential (1M, 10M, 100M, 1B), the performance grows at a slower rate. This observation suggests that there may be a limit to the amount of data that we can feed into a RoBERTa model and the knowledge that the model can acquire.

\section{Cost-benefit analysis}
\label{sec:cost_benefit}

For the sake of a more holistic view on the quality of the models, we perform a cost--benefit analysis of the performance gains in the different tasks, with an estimate of the financial and environmental cost of developing the models. As the resources used to train the MiniBERTas are not publicly available, we rely on the data provided in \citep{strubell-etal-2019-energy}
to estimate the cost of developing each individual model based on the costs of RoBERTa, trained on 30B words, in proportion to the amount of words used to train each family of models.

\paragraph{Financial cost.} As RoBERTa base was trained on 1024 Nvidia V100 GPUs for 24 hours (i.e., 24,576 GPU hours), and the price per hour of Nvidia V100 (on-demand) is \$2.48 \citep{strubell-etal-2019-energy}, the cost of training RoBERTa base amounts to \$60,948, and  the cost of training a MiniBERTas model can be estimated to be \$60,948 / 30B words * \#TrainingWords. E.g., for the \textit{roberta-base-1b} model: \$60,948 / 30B words * 1B words = \$2,032.

\begin{table*}
    \begin{tabular}{lrrrrr}
        \hline
        \thead[cl]{Model family} & \thead[cr]{Cost} & \thead[cr]{CO\textsubscript{2}e (lbs)} & \thead[cr]{PoS} & \thead[cr]{Dep. parsing} & \thead[cr]{Paraphrase id.} \\
        \hline
        roberta-base-1B	        &    \$20320	&    2330	&   96.03 (+0.5\%)	&   85.73 (+1.76\%)	    & 89.59 (+2.02\%) \\
        roberta-base-100M	    &    \$5075     &    582.5	&   95.53 (+1.11\%)	&   83.97 (+4.04\%)	    & 87.57 (+2.79\%) \\
        roberta-base-10M	    &    \$500	    &    58.25	&   94.42 (+2.73\%)	&   79.93 (+14.48\%)	& 84.78 (+5.34\%) \\
        rob-med-small-1M	&    \$50       &    5.825 	&   91.69 (base)	&   65.45 (base)	    & 79.44 (base) \\
        \hline
    \end{tabular}
    \caption{Comparison of the estimated cost of developing the different MiniBERTas families in terms of cloud compute cost (USD) and CO\textsubscript{2} emissions (lbs) and their averaged performances on PoS tagging (acc), Dep. Parsing (LAS), and Paraphrase identification (F1). In parentheses, we show the increment with respect to the previous smaller model.}\label{tab:estimated_costs}
\end{table*}

\paragraph{CO\textsubscript{2} Emissions.} Using \citet{strubell-etal-2019-energy}, we extrapolate that Nvidia V100 GPUs emit 0.28441456 lbs of CO\textsubscript{2} per GPU per hour, which means that the training of RoBERTa base emitted 6,990 lbs of CO\textsubscript{2}. We estimate the emissions of the training of each MiniBERTas model as 6,990 lbs / 30B * \#TrainingWords. \\

To develop each MiniBERTas models, \citeauthor{warstadt-etal-2020-learning} run the pretraining 10 times for the bigger family (roberta-base-1B), and 25 times for the other three families (roberta-base-100M, roberta-base-10M and roberta-med-small-1M) with varying hyperparameters. Therefore, to compute the cost of developing each family of models, we multiply the cost of training a single model by the number of pretraining runs needed to obtain it. Table \ref{tab:estimated_costs} lists the estimated costs and CO\textsubscript{2} emissions of the development of each MiniBERTas family, along with their averaged performance on the three studied downstream applications. We see that small performance gains come at high financial and environmental costs. E.g., for \textit{roberta-base-1B}, a performance increase of 0.5\%--2.02\% on downstream applications has a cost of \$20K in computing resources and significant carbon emissions, higher than the estimated 1984 lbs generated by a single passenger flying between New York and San Francisco \citep{strubell-etal-2019-energy}.

\section{Discusion and conclusions}
\label{sec:conclusions}

Our experiments shed light on the impact of pretraining data size on the syntactic capabilities of RoBERTa. Our results indicate that models pretrained with more data encode better syntactic information (as measured by \citeauthor{hewitt-manning-2019-structural}'s structural probes) and offer a higher syntactic generalization over the different syntactic phenomena covered by the tests assembled in \citep{hu-etal-2020-systematic}. Moreover, models pretrained with more data seem to be more robust to the presence of modifiers in the syntactic tests, i.e,. intervening content inserted before the critical region. As was already observed in \citep{hu-etal-2020-systematic}, there is no simple relationship between the perplexity of the models and the SyntaxGym score: the variance in intra-family SG score is not explained by the perplexity differences. When zooming in on the different test circuits, probing different linguistic phenomena, we observe that there is a high variability between the scores achieved by the models of the same family, with no single model for any family performing best across all tests. While the family pretrained with more data outperforms all the models of the other families on 3 out of 6 circuits, it offers a surprisingly low performance in Gross Syntactic State, clearly outperformed by the smaller models.

We also compare the performance of the different models fine-tuned on PoS tagging, dependency parsing and paraphrase identification, observing that models with more training data offer a better performance, and the model with the smaller architecture (roberta-med-small-1M) performs remarkably worse. However, while the amount of training data between families grows exponentially, we observe that the performance grows at a much slower rate, suggesting that there may be a limit to the knowledge that a RoBERTa model can acquire solely from raw pretraining data.

We complement our findings with a financial and environmental cost--benefit analysis of pretraining models on different amounts of data. We show that while models pretrained on more data encode more syntactic information and perform generally better on downstream applications, small performance gains come at a huge financial and environmental cost. Thus, when developing and training new models we should weigh between the benefit of making models bigger and pretraining them on huge datasets and the costs this implies, prioritizing computationally efficient hardware and algorithms.

A question that still needs to be addressed by future work is whether it is possible to complement information-theoretical metrics such as perplexity with metrics measuring specific types of knowledge, e.g., syntax, in order to develop and select more robust and efficient models to solve Natural Language Understanding tasks.

\section*{Acknowledgments}
This work has been partially funded by the European Commission via its H2020 Research Program under the contract numbers 786731, 825079, 870930, and 952133. This work has been partially supported by the ICT PhD program of Universitat Pompeu Fabra through a travel grant.

% Entries for the entire Anthology, followed by custom entries
\bibliography{anthology,custom}
\bibliographystyle{acl_natbib}

% \appendix
% 
% \section{Example Appendix}
% \label{sec:appendix}
% 
% This is an appendix.

\end{document}